\documentclass[conference]{IEEEtran}
\ifCLASSINFOpdf
   \usepackage[pdftex]{graphicx}
\else
\fi

\usepackage{amstext}

\begin{document}
%
\title{Feature Representation In Convolutional Neural Networks}

\author{\IEEEauthorblockN{Ben Athiwaratkun}
\IEEEauthorblockA{Department of Statistical Science \\Cornell University\\
Ithaca, NY 14850\\
Email: {\tt pa338@cornell.edu}}
\and
\IEEEauthorblockN{Keegan Kang}
\IEEEauthorblockA{Department of Statistical Science \\Cornell University\\
Ithaca, NY 14850\\
Email: {\tt tk528@cornell.edu}}
}

%


\maketitle

\begin{abstract}
Convolutional Neural Networks (CNNs) are  powerful  models that achieve impressive results for image classification. In addition, pre-trained CNNs are also useful for other computer vision tasks  as generic feature extractors \cite{DBLP:journals/corr/RazavianASC14}. 
This paper aims to gain  insight into the feature  aspect of CNN and demonstrate other  uses of CNN features.
Our   results show that  CNN feature maps can be used with Random Forests and SVM to yield classification results that outperforms the original CNN. A CNN that is less than optimal (e.g. not fully trained or overfitting) can also extract features for Random Forest/SVM that yield competitive classification accuracy. 
In contrast to the literature which uses the top-layer activations as feature representation of images for other tasks \cite{DBLP:journals/corr/RazavianASC14}, using lower-layer features can yield better results for classification.

\end{abstract}

\IEEEpeerreviewmaketitle

\section{Introduction}
Convolutional Neural Networks (CNNs) have proven to be very successful frameworks for image recognition. In the past few years, variants of CNN models achieve increasingly better performance on the renowned ImageNet dataset for object classification, starting from AlexNet from \cite{NIPS2012_4824}, OverFeat \cite{DBLP:journals/corr/SermanetEZMFL13}, GoogLeNet \cite{DBLP:journals/corr/SzegedyLJSRAEVR14}, and  a recent model by \cite{rectifiersdelve} with classification accuracy surpassing human-level performance. Nevertheless, there are still many aspects of CNNs that researchers are striving to understand.

In recent years, research that  seeks to  gain insights into CNN models include  exploring new non-linear activation functions, new training techniques, optimal network configurations, etc. For instance,  \cite{AISTATS2011_GlorotBB11} explores the ReLU activation function which controls sparsity and helps speed up training time, \cite{dropoutpaper} and \cite{Maxoutnetworkspaper} introduces training techniques that reduce overfitting, 
\cite{Chatfield14} explores the reduced dimensionality of CNN output layer that  yields good performance, to name but a few.
 
Collectively, these research help  increase the understanding and consequently the performance of CNNs. Our research aims to understand another aspect of CNNs  -- the feature maps.

The idea of exploring CNN features is also motivated  by their usefulness on a wide variety of tasks. 
As introduced earlier, the activations which are the output of CNN layers  can  be interpreted as  visual features.  
CNN models which are trained for classification have been used as feature extractors by removing the output layer (which output class scores). In an AlexNet, this would compute a 4096-dimensional vector for each input image (CNN codes).  In particular, a pre-trained CNN on ImageNet dataset can  be used as a generic feature extractor for other datasets \cite{DBLP:journals/corr/RazavianASC14}. Features extracted from pre-trained CNN such as OverFeat  have been successfully used in computer vision tasks such as scene recognition, object attribute detection and achieves better results compared to handcrafted features \cite{DBLP:journals/corr/RazavianASC14}. Given the usefulness of CNN features, our research aims to further assess the features  and demonstrate how we can use the features for other  tasks. 

\section{Methodology and Results}

\subsection{Model Setup}
\subsubsection{Dataset}

In our experiments, we  use the plankton data provided by Oregon State University Hatfield Marine Science Center. 

We use 30,300 labelled samples from the {\tt train} data, which comprises  121 classes. 

\begin{figure}[h!]
\begin{center}
\includegraphics[width = 2.2in]{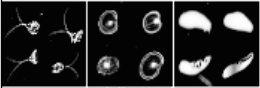}
\end{center}
\caption{Sample images of 3 classes with 4 samples shown for each class}
\label{fig:sample_classes}
\end{figure}

\subsubsection{CNN Models }
For our experiments, we use the following CNN architectures.

\begin{table}[h!]
\caption{\scriptsize Architecture of CNN 1}
\vspace*{0.1cm}
\begin{center}
{
{\scriptsize \begin{tabular}{|c|c |c | c | } \hline
{\bf Layer } & {\bf Layer Type} & {\bf Size} & {\bf Output Shape}   \\ \hline
 1 & Convolution + ReLU & 32 5$\times$5 filters  	& 	\\ \hline
1 & Max Pooling & 2$\times$2, stride 2 			& (32,12,12)		\\ \hline 
2 & Convolution + ReLU & 48 5$\times$5 filters  	&	\\ \hline
2 & Max Pooling & 2$\times$2, stride 2 			& (48,4,4) 		\\ \hline 
3 & Convolution + ReLU & 64 5$\times$5 filters  	&	\\ \hline
3 & Max Pooling & 2$\times$2, stride 2 			&  (64,1,1)		\\ \hline 
4 & Fully Connected + ReLU & 121 hidden units  	& 121	\\ \hline
5 & Softmax & 121 way 						& 121 \\ \hline
\end{tabular}
}} 
\end{center}
\label{table:CNN1}
\end{table}

\begin{table}[h!]
\caption{ \scriptsize Architecture of CNN 2}
\vspace*{0.1cm}
\begin{center}
{{\scriptsize 
\begin{tabular}{|c|c |c | c | } \hline
{\bf Layer } & {\bf Layer Type} & {\bf Size} & {\bf Output Shape}  \\ \hline
1 & Convolution + ReLU & 32 5$\times$5 filters & (32,36,36)\\ \hline
2 & Convolution + ReLU & 32 5$\times$5 filters &  \\ \hline
2 & Max Pooling & 2$\times$2, stride 2  & (32,16,16)\\ \hline 
3 & Convolution + ReLU & 48 5$\times$5 filters & (48,12,12)  \\ \hline
4 & Convolution + ReLU & 48 5$\times$5 filters & \\ \hline
4 & Max Pooling & 2$\times$2, stride 2 & (48,4,4)\\ \hline 
5 & Convolution + ReLU & 64 3$\times$3 filters  & \\ \hline
5 & Max Pooling & 2$\times$2, stride 2 & (64,1,1)\\ \hline 
6 & Fully Connected + ReLU & 121 hidden units  & 121\\ \hline
7 & Softmax & 121 way & 121 \\ \hline
\end{tabular}
}} 
\end{center}
\label{table:CNN2}
\end{table}

\begin{table}[h!]
\caption{ \scriptsize Architecture of CNN 3: Dropout used at first layer}\vspace*{0.1cm}
\begin{center}
{\scriptsize \begin{tabular}{|c|c |c | c|} \hline
{\bf Layer} & {\bf Layer Type} & {\bf Size} & Output Shape  \\ \hline
1 & Convolution + Maxout & 48 8$\times$8 filters & \\ \hline
1 & Max Pooling & 4$\times$4, stride 2 & (48,10 ,10)\\ \hline 
2 & Convolution + Maxout & 48 8$\times$8 filters  & \\ \hline
2 & Max Pooling & 4$\times$4, stride 2 & (48,4,4)\\ \hline 
3 & Convolution + Maxout & 24 5$\times$5 filters &  \\ \hline
3 & Max Pooling & 2$\times$2, stride 2 & (24,3,3)\\ \hline 
4 & Softmax & 121 way & 121\\ \hline
\end{tabular}
}
\end{center}
\label{table:CNN3}
\end{table}

We use ReLU as an activation function which is a popular choice especially for deep networks.  The activation function has been shown to speed up training time. \cite{AISTATS2011_GlorotBB11}.

We note that each sample image varies in scale and is  not necessarily square. To use these images on CNNs, we rescale them to $28 \times 28$ pixels for CNN1 and CNN3 and to $40 \times 40$ pixels for CNN2. This is because  networks with more layers (such as CNN2) generally need a larger input size since the  pooling layer exponentially reduces the size of the layer input.

\subsubsection{CNN Training }
We  follow recommended training procedures in CNN literature. We use cross-entropy loss as our objective function that we seek to minimize. We use mini-batch stochastic gradient descent with momentum which is shown to be an effective method for training CNN \cite{DBLP:journals/corr/abs-1212-0901}. We also use the max-norm constraint approach to regularize weights \cite{stanfordwebby}.
We split the samples into training, validation, and test set of size 25000, 1500, 3800 respectively.

\subsection{CNN Features for Classification}

As mentioned before, CNN models such as OverFeat, AlexNet, GoogLeNet that are pre-trained on ImageNet can been used as generic feature extractors for other tasks. This is done by removing the top output layer and using the activations from the last fully connected layer (CNN codes) as features. \cite{DBLP:journals/corr/RazavianASC14} uses pre-trained OverFeat on other datasets to extract features and use these features on other computer vision tasks. It turns out that this off-the-shelf feature extractor give features that yield better results than handcrafted features \cite{DBLP:journals/corr/RazavianASC14}.

As opposed to using CNN feature on other tasks, we are interested in using them for the original classification problem. 
To do this, we take features from our CNN trained on the plankton dataset and using them as training input for other classification methods (Random Forests and SVM). We are curious to see how the performance will compare to the baseline CNN classification accuracy.

In this research, we also take activations from other layers in addition to the last fully connected layer as feature representation of images. This is not a conventional approach in the literature \cite{DBLP:journals/corr/RazavianASC14}. We are interested to see if the lower-layer features are more suitable for classification with other algorithms.

\subsubsection{CNN vs SVM and Random Forest on CNN Features}

First, we train CNN1 described in Table~\ref{table:CNN1}. We pass the training and validation samples  to the CNN and take layer activations as  training input for SVM and Random Forest. These training and validation samples are the same samples used to train CNNs.
We use feature maps of each layer in CNN1 as training input for $3$ classification models, namely, Random Forest, SVM (one-vs-all) and SVM (one-vs-one). SVM one-vs-all trains $n$ classifiers for $n = 121$ classes. SVM one-vs-one trains ${n \choose 2}$ classifiers. 

\begin{figure}[h!]
\begin{center}
\includegraphics[width = 3.5in]{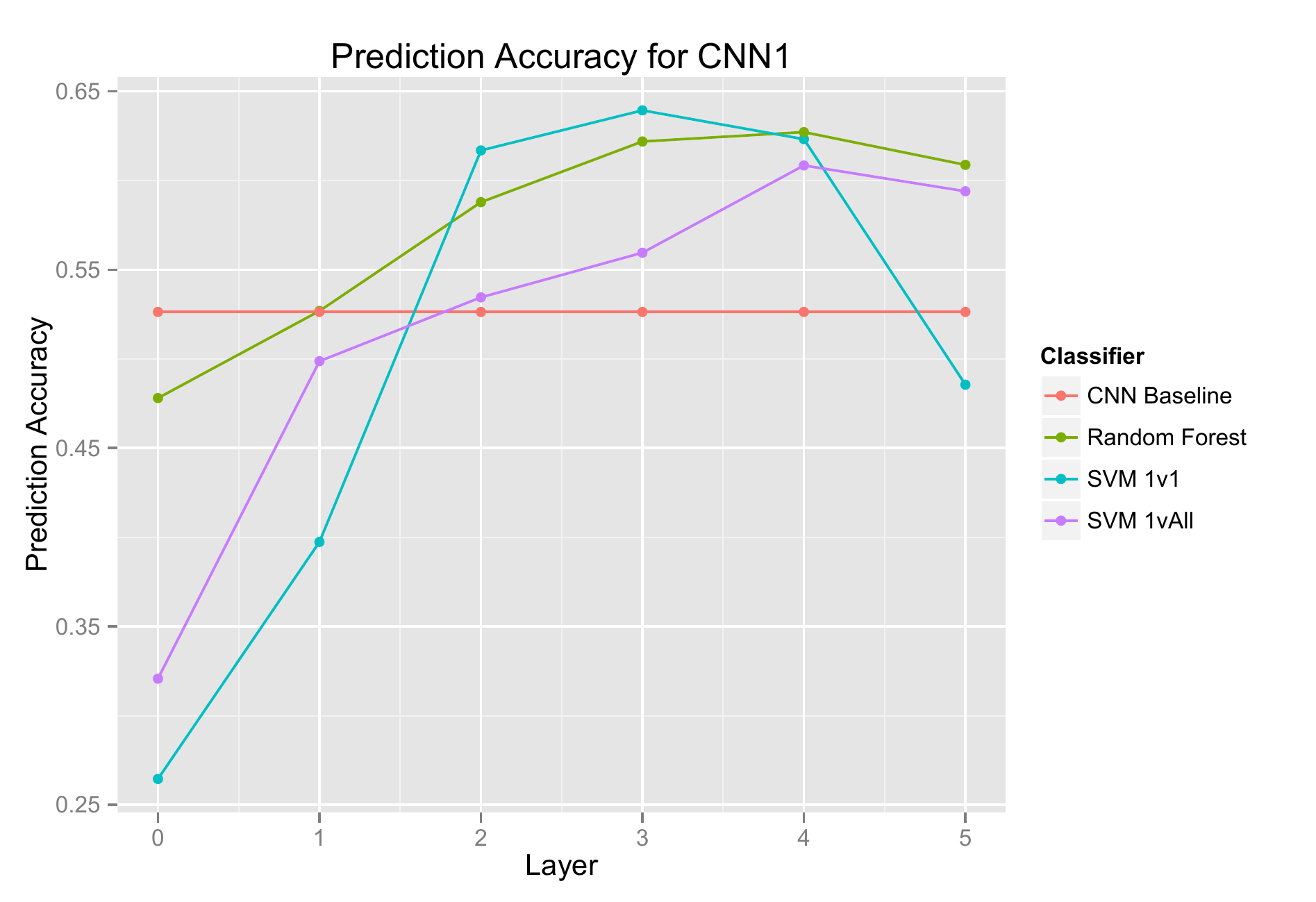}
\end{center}
\caption{Test Set Prediction Accuracy of Random Forest, SVM and baseline CNN1}
\label{fig:results_cnn1}
\end{figure}

The classification results are  shown in Figure~\ref{fig:results_cnn1}. In this figure,   layer 0  represents  using raw data (as opposed to using CNN features) for training input. This is shown as comparison for each classification model to see how well they perform without using features from CNN. The classification accuracy of CNN1 is  shown as the baseline.

A few points to highlight based on these results:
\begin{itemize}
\item Random Forest and SVM trained by CNN features can perform better than  the baseline CNN. This is quite surprising since the original CNN trains the weights which specify feature extraction. 
\item We note that the main difference between the original CNN and Random Forest/SVM on CNN features  is that  the original CNN uses the last convolutional features on the trained neural network (fully connected + output layer). This is  shown in the CNN architecture in Table~\ref{table:CNN1}. As opposed to using the trained neural network for prediction, Random Forest and SVM performs their own training by taking fixed CNN features. The features are also from many layers, not limited to the last convolutional layer. 
\item Generally, higher layers seem to extract better features as indicated by the increasing performance of SVM and Random Forest from layer 1 to layer 4. However, the highest accuracy is achieved from using layer 3  activations  on SVM one-vs-one. This confirms our hypothesis that high-level features might not necessarily be better than low-level features. We note that layer 3 corresponds to the last convolutional layer where layer 4 corresponds to the fully connected layer (see Table~\ref{table:CNN1} for more details).
\end{itemize}

\begin{figure}[h!]
\begin{center}
\includegraphics[width = 3.5in]{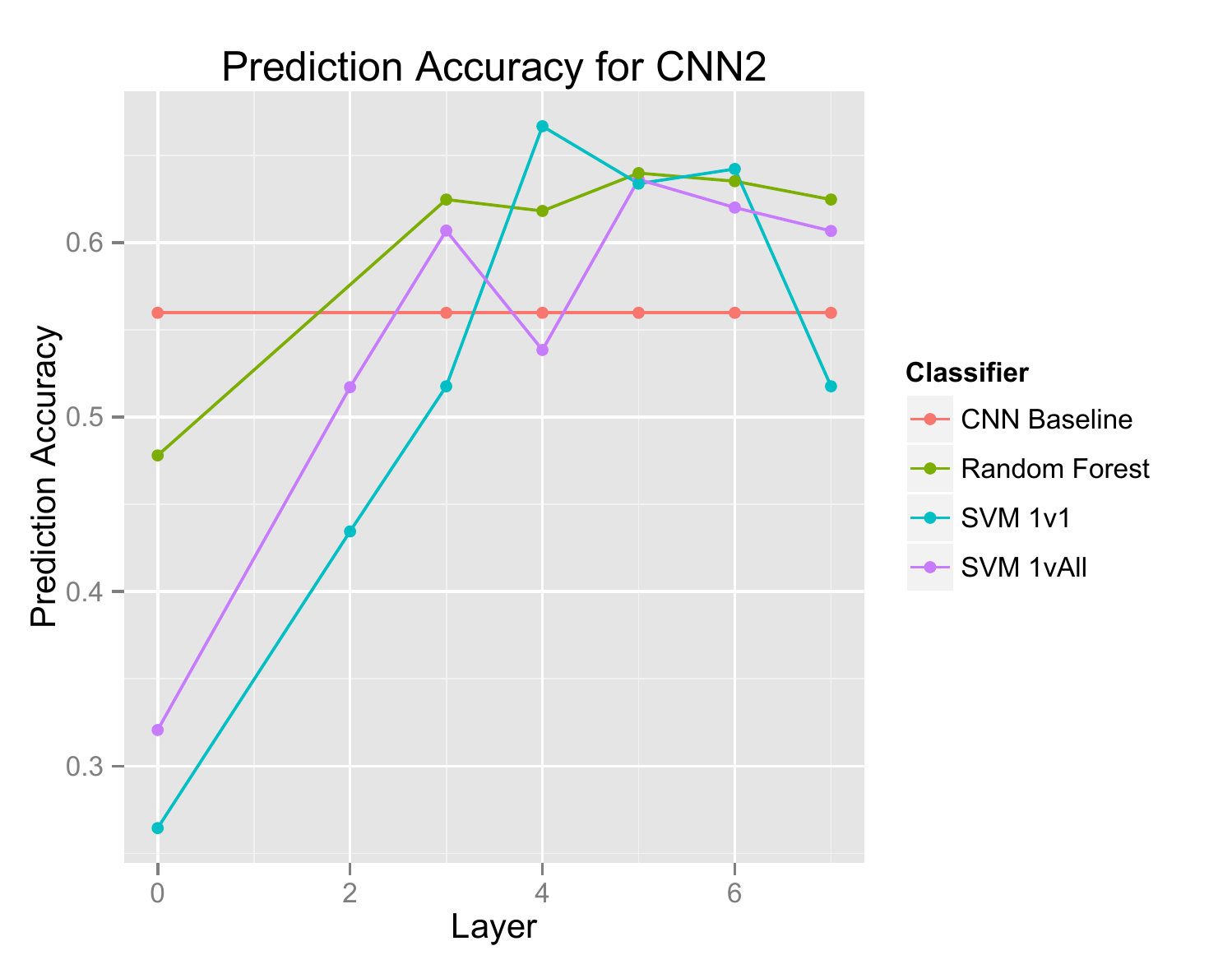}
\end{center}
\caption{Test Set Prediction Accuracy of Random Forest, SVM and baseline CNN2}
\label{fig:results_cnn2}
\end{figure}

We also did the same experiment with CNN2 (see Table~\ref{table:CNN2} for details). The results  shown in Figure~\ref{fig:results_cnn2} confirms a similar accuracy trend of increasing prediction accuracy using higher level features. However, the best prediction accuracy is achieved with SVM one-vs-one using activations from layer 4 (one layer before the last convolutional layer). We also observe that for SVM one-vs-rest and Random Forest, the best accuracy is achieved using layer 5, the last convolutional layer. This also adds to the evidence that the last fully connected features might not be the most optimal as training input for some classification models. 

\subsubsection{Feature Significance}
\begin{figure}[h!]
\begin{center}
\includegraphics[width = 3.3in]{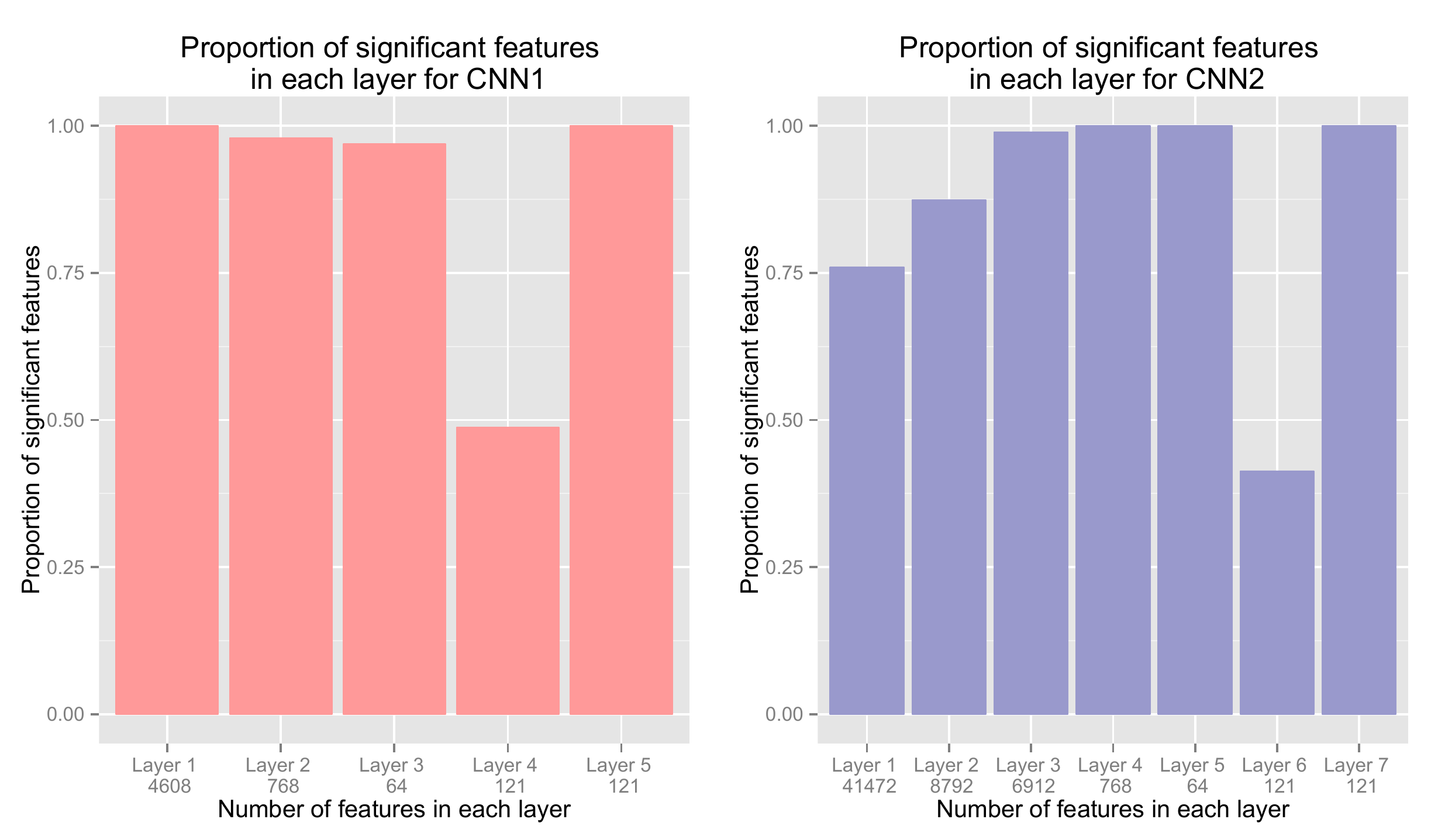}
\end{center}

\caption{CNNs Feature Importance}
\label{fig:results_significance}
\end{figure}

This section tests for significance of features with Random Forest. The test measures the relative importance of that feature based on how it influences the output prediction. Features used at the top of the tree contribute to the final prediction on a larger fraction of input samples \cite{scikit-learn}. The output of this feature importance test is a ratio from $0$ to $1$ that ranks how important the feature is. For example, if there are $n$ features and every feature is equally important, the importance scores are all $\frac{1}{n}$. To select  significant features, we impose a threshold $t = \frac{1}{100} \cdot \frac{1}{n}$ for a test with $n$ total features. Figure~\ref{fig:results_significance} shows the results for CNN1 and CNN2. At the first fully connected layer (layer 4 in CNN1 and layer 6 in CNN2),  there is a lower proportion of significant features. This might help explain why features at the fully connected layer can yield lower prediction accuracy than features at the previous convolutional layer.

\subsubsection{SVM and Random Forest on Early-Epoch CNN Features}

Based on earlier results, a CNN with low prediction accuracy can give features that yield better classification results with Random Forest and SVM. 
We thus want to see the quality of features from a CNN that is not fully trained. To do this, we extract features from CNN1 at epochs $0$ to $62$.  Note that an increment in epoch means an additional pass through the whole training data while doing stochastic gradient descent. Epoch 0 represents an initialized CNN (with randomized weights) that has not been trained at all.

\begin{figure}[h!]
\begin{center}
\includegraphics[width = 3.5in]{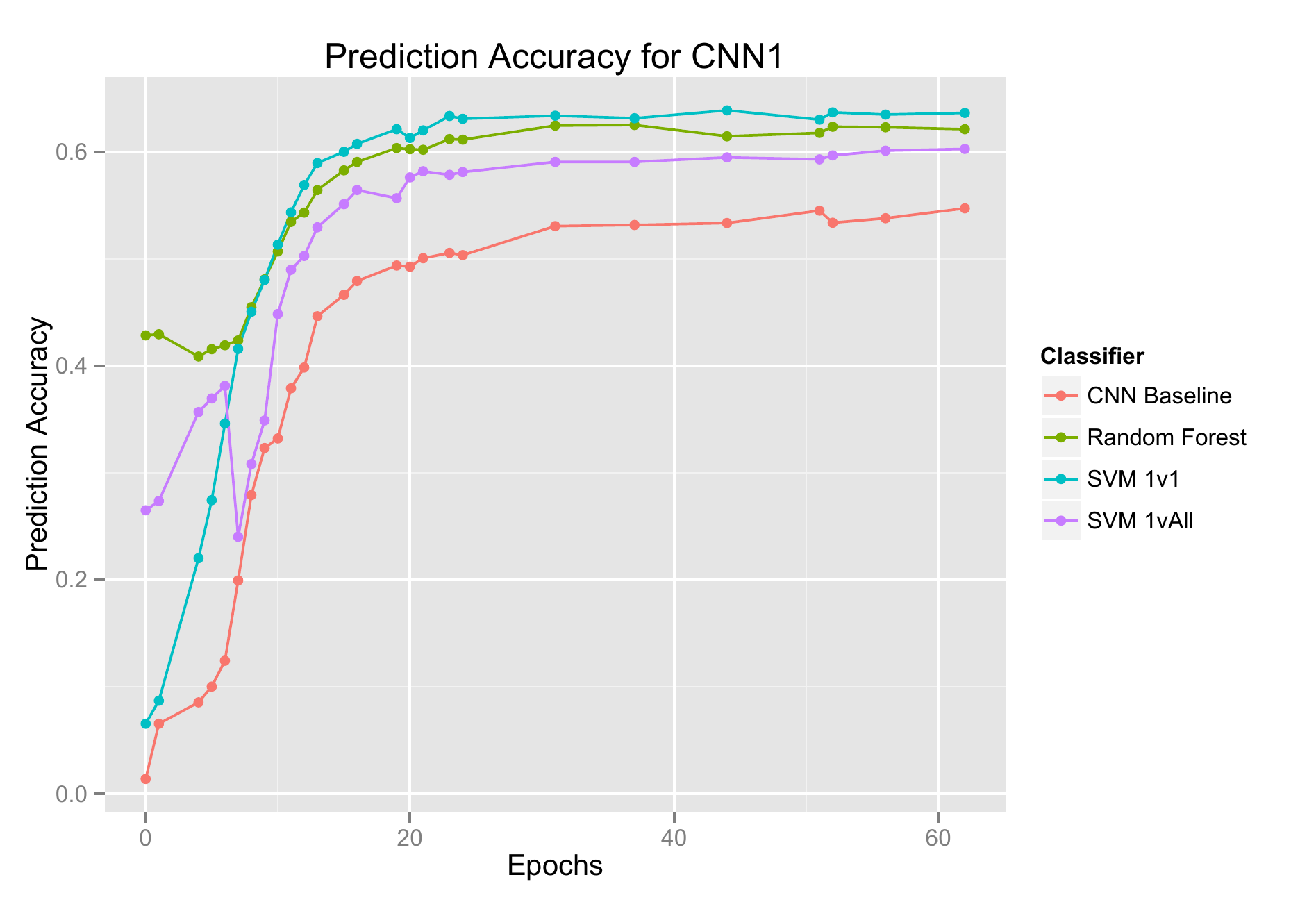}
\end{center}
\caption{Prediction accuracy of CNN, Random Forest and SVM at varying epochs. Random Forest and SVM use layer 3 activations as training input.}
\label{fig:results2}
\end{figure}

Figure~\ref{fig:results2} shows the result of this experiment. We use CNN1 with features from layer 3 as the training input for SVM and Random Forest. The trend shows that CNN generally extracts better features at increasing epochs. However, we observe that Random Forest and SVM can achieve high accuracy results even at early epoch ($24$) while CNN's accuracy is still increasing. Perhaps Random Forest and SVM can be used as to indicate an upper bound on the accuracy of CNN. Another interesting observation is that Random Forest still performs quite well on features extracted at epoch 0. These features are obtained from convolutions with weights are randomly initialized.

\subsubsection{CNN with Bagging vs Random Forest and SVM}
Bagging is an ensemble method that has been used successfully with CNNs to achieve better accuracy \cite{Ciresan:2011:FHP:2283516.2283603}. However, the performance gain is usually not dramatic. In addition, each CNN model is computationally expensive to train. Combining many CNN models takes non-trivial computational resources.  In this section, we are interested to how the bagging performance compares to that of of SVM and Random Forest trained from features extracted by one model. 

To do this, we train 8 models with CNN1 configurations. For each test sample, we obtain  class probabilities from all the trained CNNs. Then, we average the predicted probabilities and pick the class with highest probability for final prediction. We obtain a performance boost from $0.5263$ (one model accuracy) to $0.55526$. However, this is instill inferior to accuracy of Random Forest or SVM ($0.6$ or above), as shown in Figure~\ref{fig:results_cnn1}.

\subsubsection{CNN with Maxout and Dropout}
The reason for such large gap between CNN performance and SVM/Random Forest could be because of overfitting at the fully connected layer. In this experiment, we use Dropout, which  is a training technique equivalent to model averaging that improves prediction accuracy by controlling co-adaptation of weights \cite{DBLP:journals/corr/abs-1207-0580}. Maxout is also an activation function that can be used with Dropout to further improves the accuracy \cite{Maxoutnetworkspaper}. We train the CNN3 model (Table~\ref{table:CNN3}) that uses Maxout and Dropout. Similar to the previous section, we also extract the features as training input for Random Forest and SVM.
\begin{table}[h!]
\caption{ \scriptsize Accuracy of Random Forest and SVM on last layer CNN features}
\vspace*{0.1cm}
\begin{center}
{{\scriptsize 
\begin{tabular}{|c|c | } \hline
{\bf Classifier } & {\bf Accuracy}  \\ \hline
Baseline CNN	&	0.64052		\\ \hline
Random Forest	&	0.65526	\\ \hline
SVM One versus Rest	&	 0.65368	\\ \hline
SVM One versus One	&	0.57184	\\ \hline
\end{tabular}
}} 
\end{center}
\label{table:CNNm}
\end{table}

Table~\ref{table:CNNm} shows that the baseline CNN3 with Maxout and Dropout having better accuracy ($0.64052$) compared to the original models ($0.5263$ for CNN1 and $0.5597$ for CNN2). This is not surprising since Maxout and Dropout have been shown to improve accuracy results on many CNN models. However, we still obtain higher accuracy than the CNN baseline by using Random Forest and SVM on the last layer features of CNN3.

We note that a model with dropout  takes about $\approx 3$ times longer to train than the original model. This is also known in  literature \cite{dropoutpaper}. Random Forest and SVM on CNN1 features  yield accuracy up to $0.6393$ (see Figure~\ref{fig:results_cnn1})  which is a competitive result without using much additional computational resources.

\subsection{CNN Features for Clustering}

In this section, we demonstrate another use for CNN features for clustering and qualitatively explain why CNN features work well. This adds to the evidence of CNN as a good feature extractor. 

We consider the task of clustering $121$ classes of plankton based on visual similarity.  A naive approach for clustering is to find the centroids of each class in our original image space, and use a hierarchical clustering algorithm with the Euclidean distance as a distance metric. This performs poorly, as there are classes where samples look different from each other (e.g. under rotation -- see Figure~\ref{fig:sample_classes} ). Thus the centroid of such classes would be a blob which gives little information about that class. 

We thus propose using the features extracted at a convolutional layer of a CNN for clustering purposes. We extract the features at the third layer of CNN1, calculate the centroid of each class in the feature space, and use a hierarchical agglomerative clustering algorithm with the Euclidean distance to cluster plankton, forming a dendrogram.

\begin{figure}[here!]
\begin{center}
\includegraphics[width=2.0in]{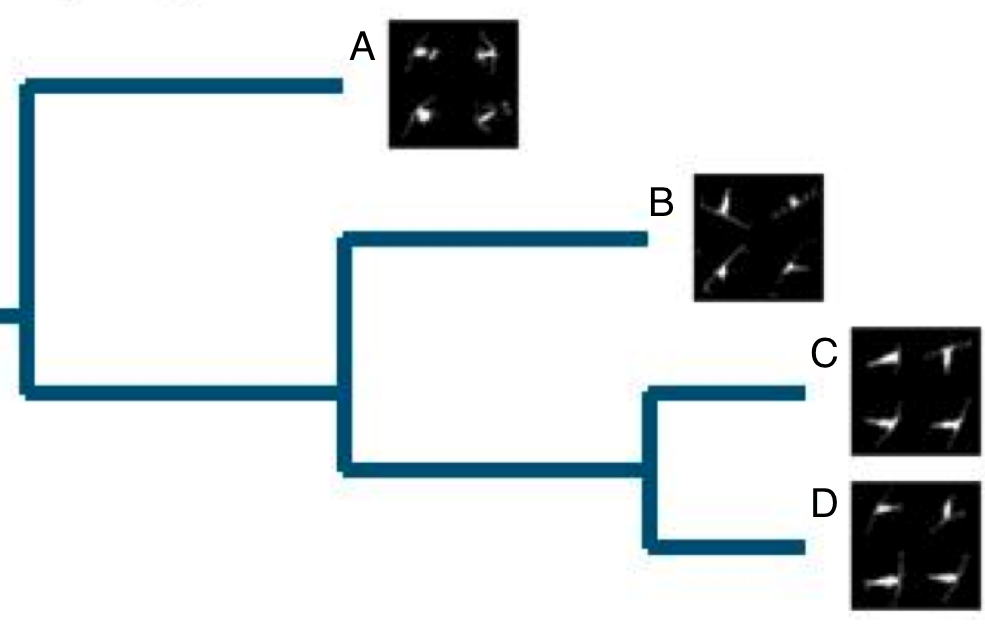}
\caption{Part of dendrogram showing similar classes}
\label{zoominfig}
\end{center}
\end{figure}

\begin{figure}[here!]
$$
\begin{array}{c}
\begin{array}{c c}
\includegraphics[trim= 0 0 0 0,width = 2.5in]{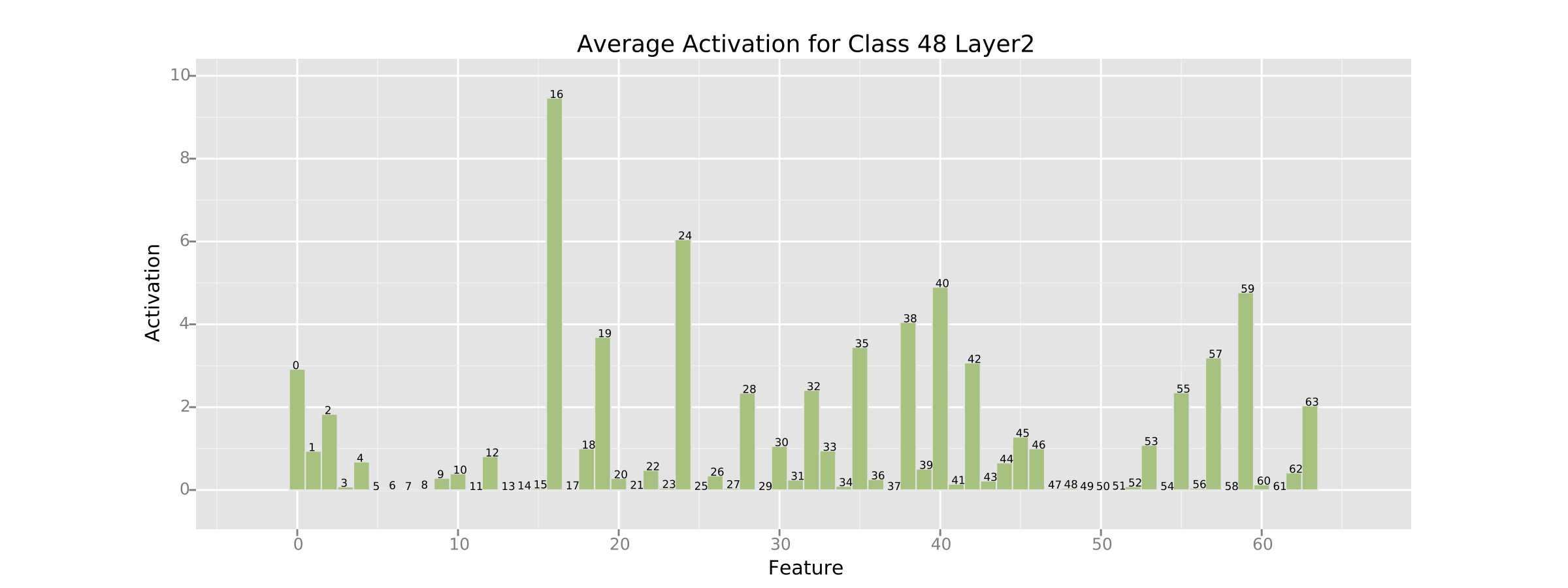} & \includegraphics[width = 0.7in]{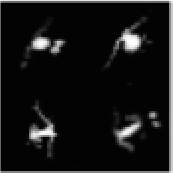}
\end{array} \\
\text{Class A: Copepod Calanoid Octomoms}\\
\begin{array}{c c}
\includegraphics[width = 2.5in]{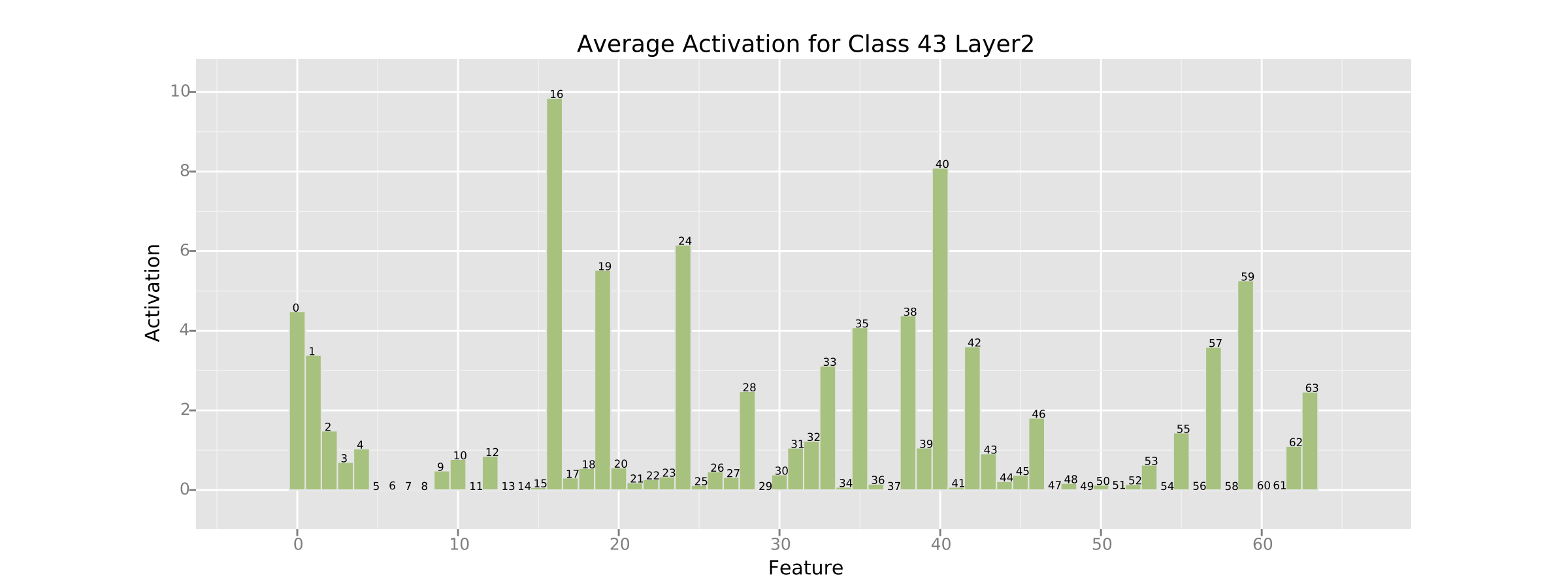} & \includegraphics[width = 0.7in]{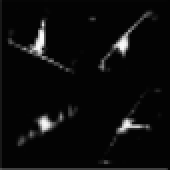}
\end{array} \\
\text{Class B: Copepod Calanoid Frilly Antennae}\\
\begin{array}{c c}
\includegraphics[width = 2.5in]{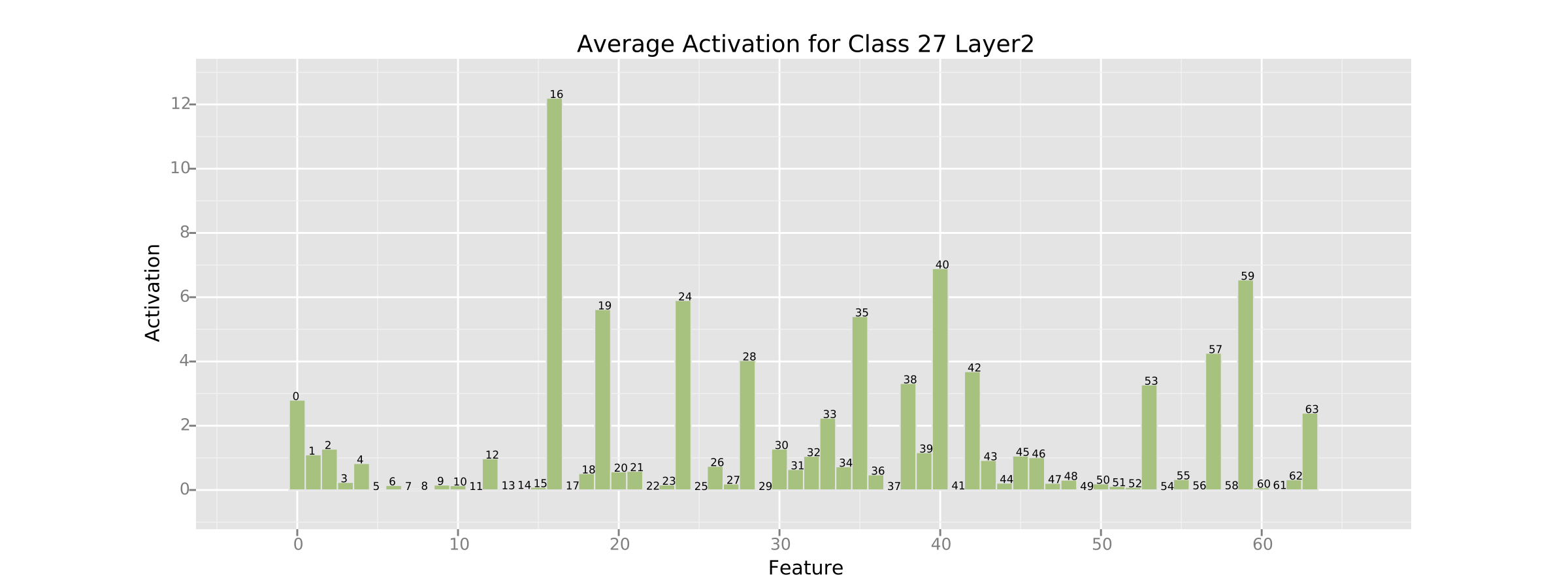} & \includegraphics[width = 0.7in]{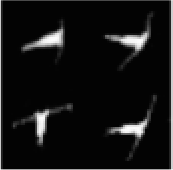}
\end{array} \\
\text{Class C: Copepod Calanoid Flatheads} \\
\begin{array}{c c}
\includegraphics[width = 2.5in]{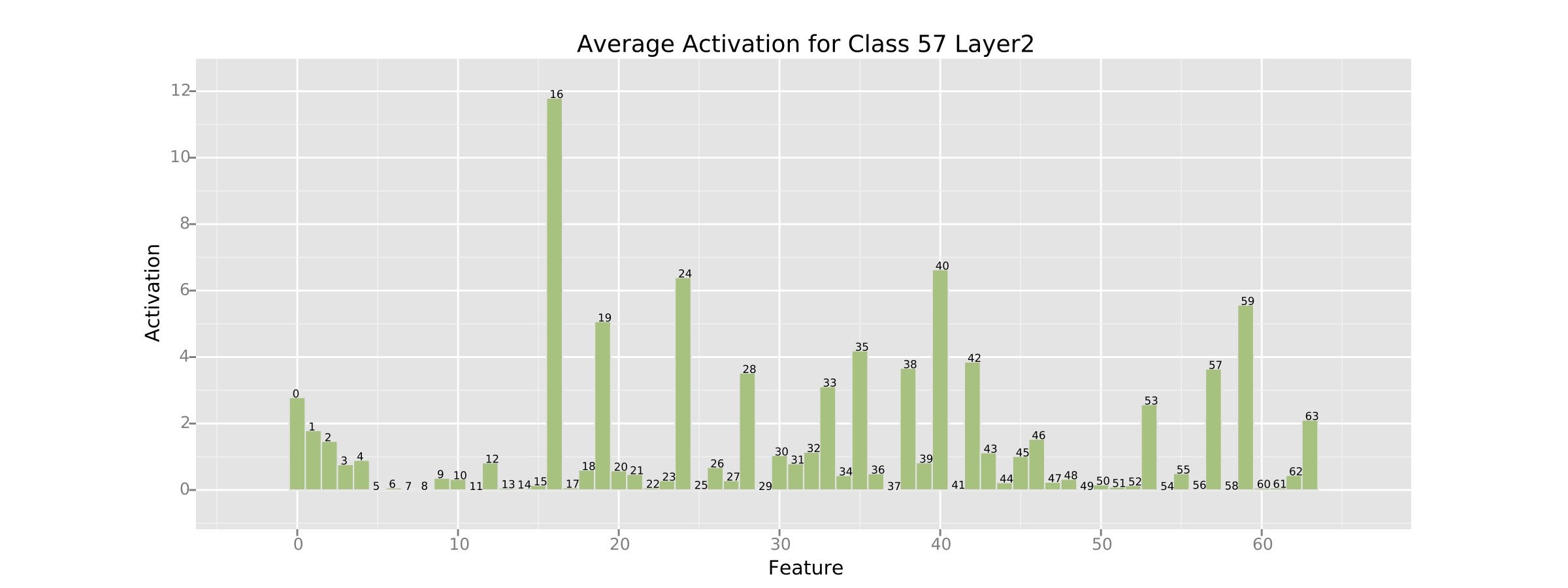} & \includegraphics[width = 0.7in]{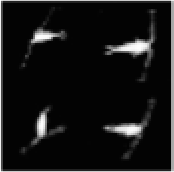}
\end{array} \\
\text{Class D: Copepod Calanoid} 
\end{array}
$$
\caption{Class Samples and Feature Scores}
\label{copepod}
\end{figure}

We look at a part of the dendrogram  in  {Figure~\ref{zoominfig}} and give some reasoning as to why the 4 classes shown are clustered closely together. Figure~\ref{copepod} show class centroids (64 dimensional vectors) which represent average feature scores for the respective classes.  Based on this figure, the top 3 common scores correspond to features 16, 24, and 40. 



\begin{figure}[here!]
\begin{center}
\includegraphics[width = 2.5in]{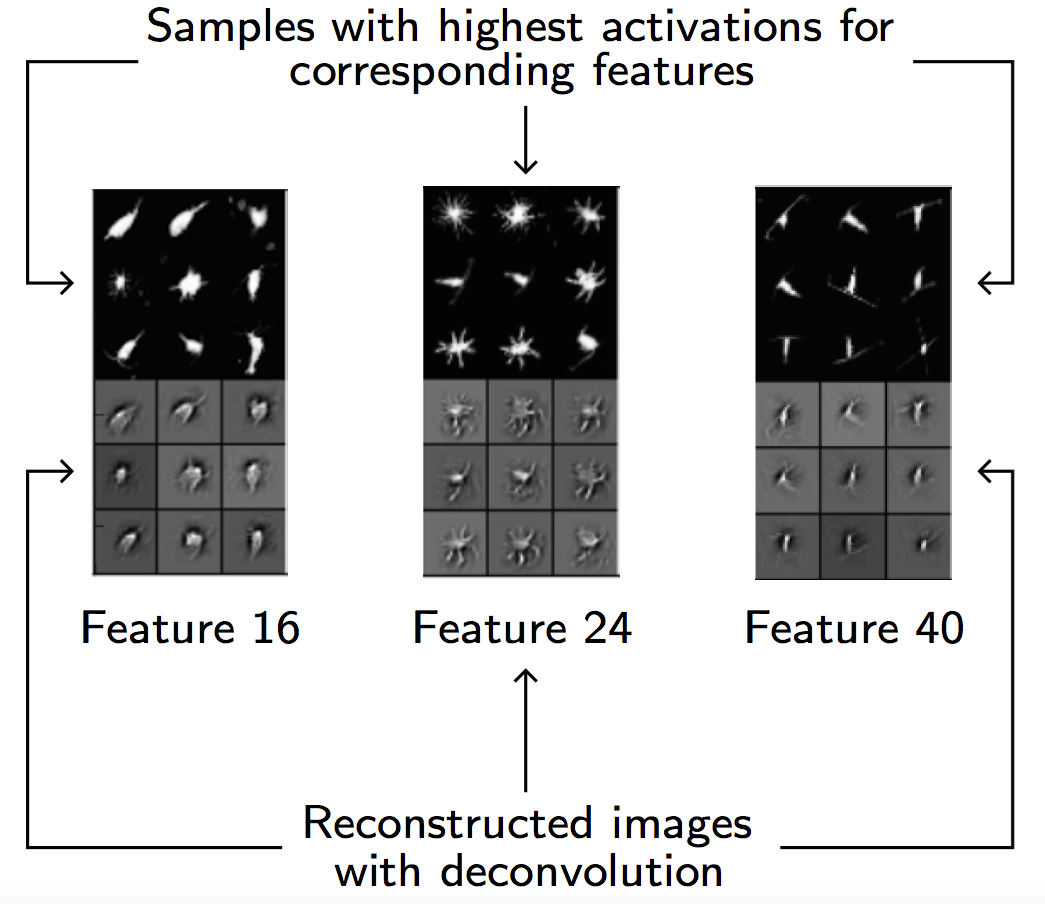} \\
\caption{Visualization of Features}
\label{fig:feats}
\end{center}
\end{figure}

To show what feature 16, 24, and 40 represent, we use a visualization technique (DeConvnet) \cite{deconv} which show parts of images that most activate the corresponding features. The visualization in Figure~\ref{fig:feats} shows that the feature 16 could represent ``rounded blobs", and feature 24, 40 could be ``small tentacles", which are common to these 4 classes. Since the CNN does clustering based on these features, this helps to explain why we get good clustering results.

{{Figure~\ref{zoominfig2} is another part of the dendrogram, which shows another group of planktons. We can see based on Figure~\ref{copepod} and ~\ref{class_samples2} that the two groups which are far apart in the dendrogram have very different feature scores. However, within each group, the feature scores are strikingly similar. In Figure~\ref{class_samples2}, we can see that features 5 and 36 have high activations in these 3 classes. Based on feature visualization,  it looks like feature 5 could be ``membranes", but feature 36 just ``porous body". 

\begin{figure}[here!]
\begin{center}
\includegraphics[width = 1.8in]{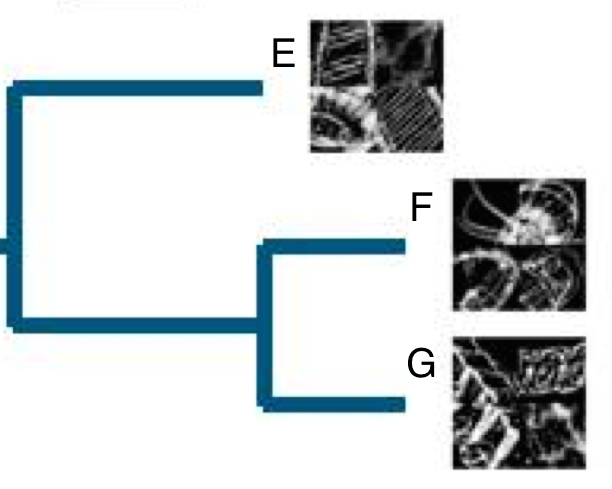}
\caption{Part of dendrogram showing similar classes}
\label{zoominfig2}
\end{center}
\end{figure} 

\begin{figure}[here!]
\begin{center}
$$
\begin{array}{c}
\begin{array}{c c}
\includegraphics[width = 2.2in]{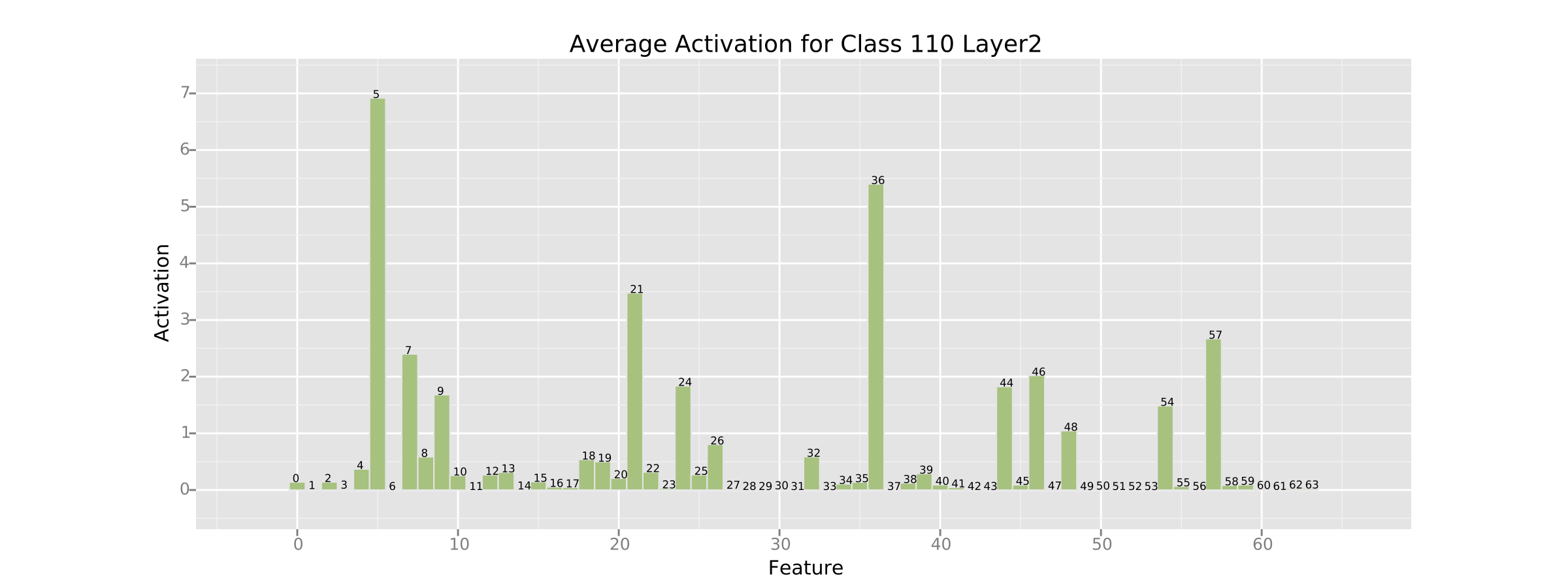} & \includegraphics[width = 0.7in]{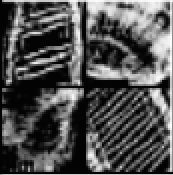}
\end{array} \\
\text{Class E: Tunicate Partial}
\end{array}
$$
$$
\begin{array}{c}
\begin{array}{c c}
\includegraphics[width = 2.2in]{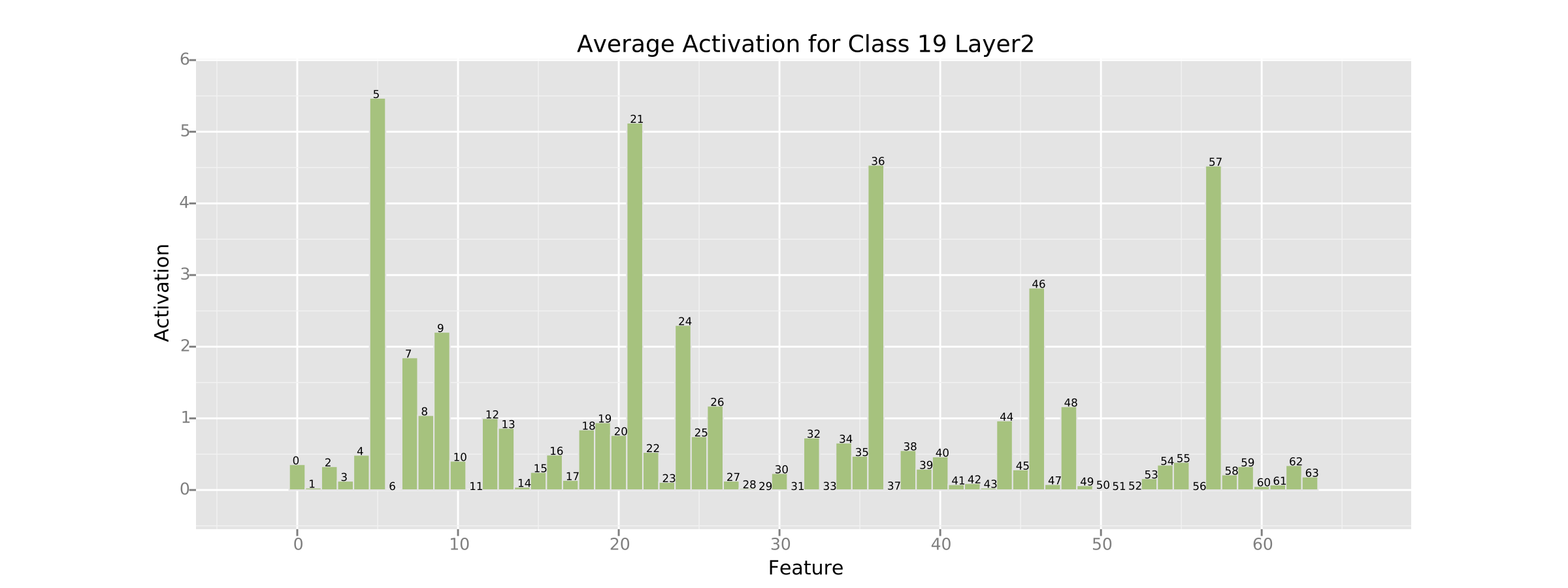} & \includegraphics[width = 0.7in]{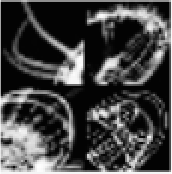}
\end{array} \\
\text{Class F: Hydromedusae Partial Dark}
\end{array}
$$
$$
\begin{array}{c}
\begin{array}{c c}
\includegraphics[width = 2.2in]{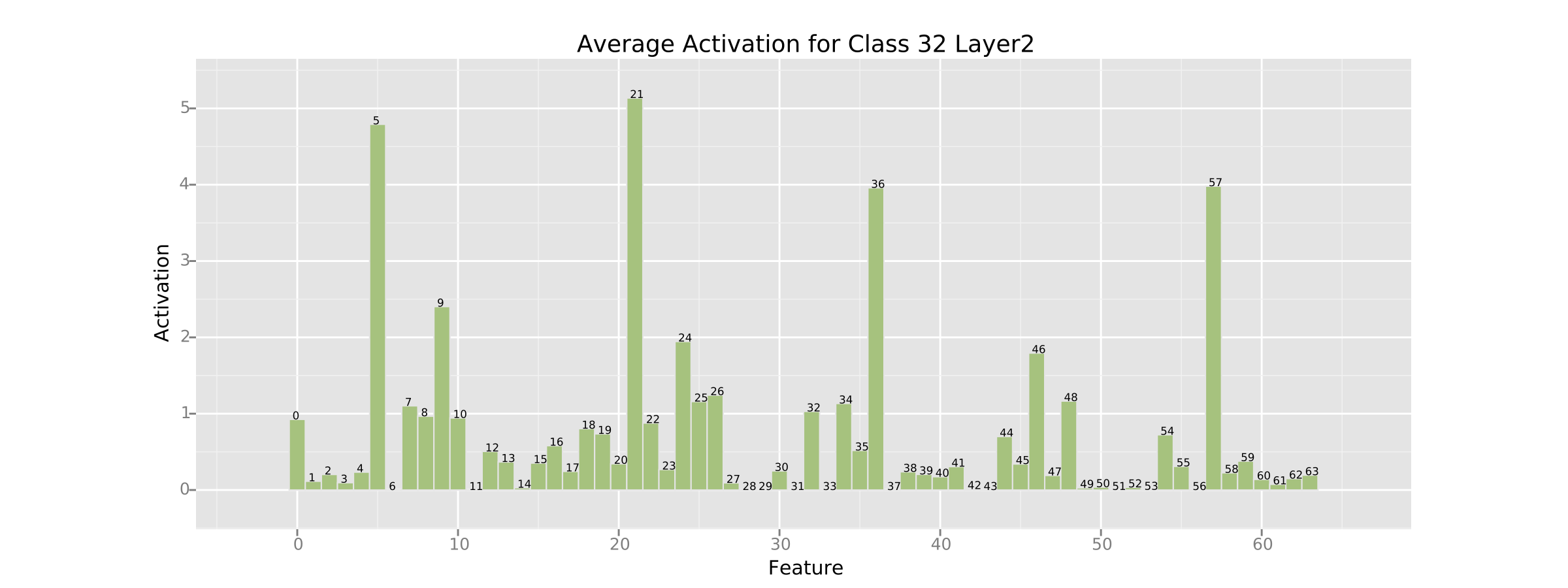} & \includegraphics[width = 0.7in]{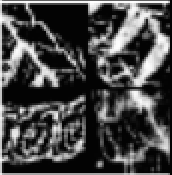}
\end{array} \\
\text{Class G: Siphonophore Partial}
\end{array}
$$
$$
\begin{array}{c c c}
\begin{array}{c}
\includegraphics[width = 0.6in]{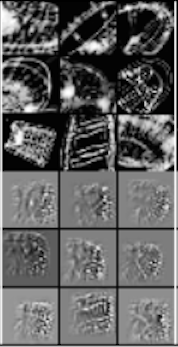} \\
\text{Feature 5}
\end{array} & \hspace{1cm}  \begin{array}{c}
\includegraphics[width = 0.6in]{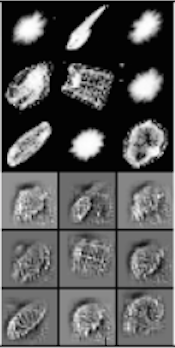} \\
\text{Feature 36}
\end{array}
\end{array}
$$
\caption{Classes Samples and Feature Vectors}
\label{class_samples2}
\end{center}
\end{figure}


We believe that this can help in the development of phenetics, which is a method of classifying organisms from a species based on their visual similarity. Thus when biologists use a CNN for image classification, they can also extract the features and get such a  dendrogram ``for free", which could be used as a dichotomous key.} 
We also note that the features involving in the distance can also be picked according to feature visualization (DeConvNet). This would give a different metric which could potentially be useful for generating clusters based on different feature qualities.

\section{Conclusion}

Our results show that Random Forest and SVM can be used with features from CNN to yield better a prediction accuracy compared to the original CNN. Even if the CNN is not optimal, e.g. not fully trained or overfits, it can still extract good features that give competitive prediction accuracy against more computationally expensive methods such as model averaging or CNN trained with Dropout. 



In addition, we found that in contrast to the practice  of using the features from the last fully connected layer (CNN codes), using lower-level features can be more optimal, at least for classification with SVM and Random Forest. 

Our qualitative analysis also  demonstrates why  CNN features are useful in  computer vision tasks. Instead of viewing the CNN as a black box, the visualization technique DeConvNet helps explains how similar images have similar CNN features. 


\section{Future Work}
We note that there are other CNN architectures that yield better classification accuracy for this dataset. Future work includes replicating these CNN architectures and use the CNN features with Random Forest or SVM. Our research can also be extended to study  pre-trained CNNs such as  AlexNet, GoogLeNet, etc. It would be ideal to experiment whether these architectures share the same trend we found that the CNN features can yield better classification accuracy compared to the original CNN. We would also like to see if the convolutional layers will yield better classification accuracy compared to the last fully connected layer which is used as traditional CNN features in literature. 

We can also experiment on other datasets such as CIFAR-10, CIFAR-100, MNIST, and ImageNet, etc. However, for large datasets such as ImageNet with over $1$ million training samples,  Random Forest might not be able to scale well. In our experiment, training Random Forest takes a significant amount of memory with only $25K$ training samples on $400$ trees. In addition, for dataset with large number of classes such as ImageNet  ($1000$ classes), the one-vs-one SVM which trains ${n \choose 2}$ classifiers for $n$ classes might be too slow as well. However, SVM (one-vs-all) which trains $n$ classifiers  should be able to handle this.  




\section{Acknowledgement}

We are grateful for the availability of the following code that is made public, as well as packages used in our research.

\begin{itemize}
\item [$\bullet$] We use {\tt pylearn2} \cite{pylearn2_arxiv_2013}   to build and train GPU-accelerated CNNs  and  {\tt theano} \cite{bergstra+al:2010-scipy} \cite{Bastien-Theano-2012}  for other CNN-related GPU computations. 
\item [$\bullet$] We adapted Deconvolutional Network code from {\tt https://github.com/ChienliMa/DeConvNet} which is built for a CNN architecture on CIFAR-10 dataset.
\item [$\bullet$] We use {\tt scikit-learn} for SVM,  Random Forest, and significant feature test.
\end{itemize}




%
\bibliography{database}
\bibliographystyle{IEEEtran}
\end{document}